\documentclass{nature}
\usepackage{amssymb,amsfonts,amsmath}
\usepackage{graphicx} 
\usepackage{gensymb}
\usepackage[normalem]{ulem}

 \usepackage{multirow}
\usepackage{epsfig}
\usepackage[figoff]{figcaps}

\usepackage{caption}
\usepackage[skip=2pt]{caption}
\usepackage{outline}
\usepackage{pmgraph}
\usepackage[normalem]{ulem}
\usepackage[utf8]{inputenc}
\usepackage{amssymb}
\usepackage{hyperref}
\usepackage{amsmath}
\usepackage{graphicx}
\usepackage{times}
\usepackage{xcolor}
\usepackage{xspace}
\usepackage[colorinlistoftodos]{todonotes} % for \todo
\usepackage{cite}
\usepackage{bm} % bold
\usepackage{url}

\usepackage{epstopdf}
\AppendGraphicsExtensions{.tiff}
\graphicspath{{fig/}} % Directory in which figures are stored

\usepackage{epsfig}
\usepackage{tikz}
\usetikzlibrary{spy}
\usepackage{algpseudocode}
\usepackage{algorithm}
\usepackage{mathrsfs}

\usepackage{amsmath,graphicx,caption,mathtools,amssymb,amsthm}
\usepackage{graphicx}
\usepackage{multirow}
\usepackage{subcaption}
\usepackage{accents}
\newcommand{\add}[1] {\textcolor{blue}{#1}} % for revision
\def\QED{~\rule[-1pt]{5pt}{5pt}\par\medskip}

\long\def\comment#1{} % comment out text

\newcommand{\xmath}[1] {\ensuremath{#1}\xspace}
\newcommand{\blmath}[1] {\xmath{\bm{#1}}}

\newcommand{\fb}{{\blmath f}}
\newcommand{\gb}{{\blmath g}}

\newcommand{\vb}{{\blmath v}}

\newcommand{\xb}{{\blmath x}}
\newcommand{\yb}{{\blmath y}}

\newcommand{\Xc}{\mathcal{X}}
\newcommand{\Yc}{\mathcal{Y}}

\newcommand{\beq}{\begin{equation}}
\newcommand{\eeq}{\end{equation}}
\newcommand{\beqa}{\begin{eqnarray}}
\newcommand{\eeqa}{\end{eqnarray}}

\usepackage{url}
\usepackage{hyperref} 

\renewcommand{\add}[1] {\textcolor{black}{#1}} % for revision

\usepackage{lineno}

\usepackage{graphicx}
\makeatletter
\let\saved@includegraphics\includegraphics
\AtBeginDocument{\let\includegraphics\saved@includegraphics}
\renewenvironment*{figure}{\@float{figure}}{\end@float}
\makeatother

\title {Deep learning enables reference-free isotropic super-resolution for volumetric fluorescence microscopy}
\author{Hyoungjun Park$^{*,1}$, Myeongsu Na$^{2}$, Bumju Kim$^{4}$, Soohyun Park$^{4}$, Ki Hean Kim$^{4,5}$, Sunghoe Chang$^{2,3}$, and Jong Chul Ye$^{\dagger,1}$}

\begin{document}
\maketitle

\begin{affiliations}
\linespread{1.2}
\baselineskip 0.23in
\item[1:] Department of Bio and Brain Engineering, KAIST, Daejeon, Republic of Korea
\item[2:] Department of Physiology and Biomedical Sciences, Seoul National University College of Medicine, Seoul, Republic of Korea
\item[3:] Neuroscience Research Institute, Seoul National University College of Medicine, Seoul, Republic of Korea
\item[4:] Department of Mechanical Engineering, Pohang University of Science and Technology, Pohang, Republic of Korea
\item[5:] Division of Integrative Biosciences and Biotechnology, Pohang University of Science and Technology, Pohang, Republic of Korea
\end{affiliations}

%\setstretch{2}
\begin{abstract}
\baselineskip 0.23in

Volumetric imaging by fluorescence microscopy is often limited by anisotropic spatial resolution from inferior axial resolution compared to the
lateral resolution. To address this problem, here we present a deep-learning-enabled unsupervised super-resolution technique that enhances anisotropic images in volumetric fluorescence microscopy. In contrast to the existing deep learning approaches that require matched high-resolution target volume images, our method greatly reduces the effort to put into practice as
 the training of a network requires as little as a single 3D image stack, without a priori knowledge of the image formation process, registration of training data, or separate acquisition of target data.
 This is achieved based on the optimal transport
driven cycle-consistent generative adversarial network that learns from an unpaired matching between high-resolution 2D images in lateral image plane and low-resolution 2D images in the other planes. 
 Using fluorescence confocal microscopy and light-sheet microscopy, we demonstrate that the trained network not only enhances axial resolution, but also restores suppressed visual details between the imaging planes and removes imaging artifacts.

\end{abstract}
 
\newpage

%High-throughput imaging (HTI, see Glossary) refers to the use of automated microscopy and image analysis to visualize and quantitatively capture cellular features at a large scale. 
Three-dimensional (3D) fluorescence imaging reveals important structural information about a biological sample that is typically unobtainable from a two-dimensional (2D) image. The recent advancements in tissue-clearing methods\cite{Chung2013,Chung2013clarity,Yang2014,Richardson2015,Hama2015}, and light-sheet fluorescence microscopy (LSFM)\cite{Santi2011, Huisken2004, Keller2008, Verveer2007} have enabled streamlined 3D visualization of a biological tissue at an unprecedented scale and speed, sometimes even in finer details. 
Nonetheless, spatial resolution of 3D fluorescence microscopy images is still far from perfection: isotropic resolution remains difficult to achieve. 

Anisotropy in fluorescence microscopy typically refers to more blurriness in the axial imaging plane. Such spatial imbalance in resolution can be attributed to many factors including diffraction of light, axial undersampling, and degree of aberration correction. Even for super-resolution microscopy\cite{Hell2009}, which in essence surpasses the light diffraction limits, such as 3D-structural illumination microscopy (3D-SIM)\cite{Gustafsson2000, Schermelleh2008} or stimulated emission depletion (STED) microscopy\cite{Hell1994}, matching axial resolution to lateral resolution remains a challenge\cite{Schermelleh2010}. While LSFM, where a fluorescence-excitation path does not necessarily align with a detection path, provides substantial enhancement to axial resolution\cite{Verveer2007}, a truly isotropic point spread function (PSF) is difficult to achieve for most contemporary light-sheet microscopy techniques, and axial resolution is usually 2 or 3 times worse than lateral resolution\cite{Power2017,ahrens2013whole,glaser2017light}. %\note{Needs backgrounds on CFM as well...}

In the recent years of image restoration applied in fluorescence microscopy, deep learning emerged as an alternative, data-driven approach to replace classical deconvolution algorithms. Deep learning has its advantage in capturing statistical complexity of an image mapping and enabling end-to-end image transformation without painstakingly fine-tuning parameters by hand. Some examples include improving resolution across different imaging modalities or numerical aperture sizes\cite{Wang2019}, or towards isotropy\cite{Weigert2018, Weigert2017} or less noise\cite{Weigert2018}. While these methods provide some level of flexibility in practical operation of microscopy, these deep-learning-based methods must assume knowledge of a target data domain for the network training: for example, high-resolution reference images as ground-truths\cite{Wang2019} or the 3D structure of the physical PSF as a prior\cite{Weigert2018, Weigert2017, Zhang2019}. Such assumption requires the success of image restoration to rely on the accuracy of priors, and adds another layer of operation to microscopists. Especially for high-throughput volumetric fluorescence imaging, imaging conditions are often subject to fluctuation, and visual characteristics of samples are considered diverse. Consequently, uniform assumption of prior information throughout a large-scale image volume could result in over-fitting of the trained model and exacerbate the performance  and the reliability of image restoration. 

In light of this challenge, the recent approach of unsupervised learning using cycle-consistent generative adversarial network (cycleGAN)\cite{Zhu2017}
is a promising direction for narrowing down the solution space for ill-posed inverse problems in optics\cite{Sim2020, Lim2020}. Specifically, it is advantageous in practice as it does not require matched pairs of data for training. When formulated as optimal transport problem between two probability distributions\cite{villani2008optimal,Sim2020}, unsupervised learning-based deconvolution microscopy can successfully transport the distributions of blurred images to high-resolution microscopy images by estimating the blurring PSF and deconvolving with it\cite{Lim2020}. Moreover, if the structure of the PSF is partially or completely known, one of the generator could be replaced by a simple operation,
which significantly reduces the complexity of the cycleGAN and makes the training  more stable \cite{Lim2020}. 
Nonetheless, one of the remaining technical issues is the difficulty of obtaining additional volumes of  high resolution microscopy images under similar experimental conditions, such as noises, illumination conditions, etc, so that they can be used as unmatched target distribution for the optimal transport. In particular, obtaining such a reference training data set with 3D isotropic resolution is yet challenging in practice.

% we propose that learning image transformations in an unsupervised manner that preserves their reversibility is a good approach to capture the statistical complexity of large-scale images, while providing flexibility and convenience of operation in practice.

To address this problem, here we present a novel unsupervised deep learning framework that blindly enhances the axial resolution in fluorescence microscopy, given a single 3D input image volume. The network can be trained with as few as one image stack that has anisotropic spatial resolution without requiring high resolution isotropic 3D reference volumes. Thereby, the need to acquire additional training data set under similar experimental condition is completely avoided. Our framework takes advantage of forming abstract representations of imaged objects that are imaged coherently in lateral and axial views: for example, coherent perspectives of neurons, where there are enough 2D snapshots of a neuron to reconstruct a generalized 3D neuron appearance. Then, our unsupervised learning scheme uses the abstract representation to decouple only the resolution-relevant information from the images, as well as undersampled or blurred details in axial images. For the complete theoretical background, refer to the Methods.

The overall architecture of the framework is inspired by the optimal transport driven cycle-consistent generative adversarial networks (OT-cycleGAN) \cite{Sim2020}. Figure \ref{fig:snu_cfm}d illustrates the learning scheme of the framework (more details are available in Supplementary Figure \ref{fig:framework_overview}). We employ two 3D generative networks ($G$ and $F$ in Fig. \ref{fig:snu_cfm}d) that respectively learn to generate an isotropic image volume from an anisotropic image volume (the forward or super-resolving path) and vice versa (the backward or blurring path). To curb the generative process of these networks, we employ two groups of 2D discriminative networks ($D_X$ and $D_Y$ in Fig. \ref{fig:snu_cfm}d). Our key innovation comes from an effective orchestration of the networks' learning based on what we feed the discriminitive networks with during the learning phase. In the forward path, the discriminative networks of $D_X$ compare the 2D axial images from the generated 3D image volume to the 2D lateral images from the real 3D image volume, while preserving the lateral image information. This pairing encourages 3D generative network $G$  to enhance only the axial resolution in the 3D volume output. On the other hand, the discriminative networks of $D_Y$ in the backward path compare the 2D images from the reconstructed 3D image volume to the 2D images from the real 3D image volume in each corresponding orthogonal plane: thereby, 3D generative network $F$ learns to revert the volume restoration process. The cycle-consistency-loss stabilizes the learning process and guides $G$ and $F$ to being mutually inverse. %Eventually, if the super-resolution is correctly learned, the image distribution in the latent space from the generated axial images will be indistinguishable from its counterpart from the real lateral images. 
By achieving the balance of loss convergence in the form of a mini-max game\cite{Goodfellow2014} by this ensemble of discriminative networks and generative networks, the network is trained to learn the transformation from the original anisotropic resolution to the desired isotropic resolution.

We demonstrated the success of the framework in confocal flouorescence microscopy (CFM) and open-top light-sheet microscopy (OT-LSM). In the CFM case, we addressed anisotropy that is mainly driven by light diffraction and axial-undersampling. We compared the results to the image volume that is imaged at a perpendicular angle. In the OT-LSM case, we address anisotropy that is driven by optical aberration caused by a refractive index mismatch and also investigated the PSF deconvolution capability of our method. In both cases, our reference-free deep-learning-based super-resolution approach was effective at improving the axial resolution, while preserving the information in the lateral plane and also restoring the suppressed microstructures.

\section*{Results}

% Visual points
% anatomical accuracy
% texture 
% interpolated details 

\textbf{Resolution enhancement in confocal fluorescence microscopy} We initially demonstrated the resolution improvement in axial plane by imaging a cortical region of a Thy 1-eYFP mouse brain with CFM. The sample was tissue-cleared and was imaged in 3D using optical sectioning. The optical sectioning by CFM is set up so that the image volume exhibits a stark contrast between lateral resolution and axial resolution, with estimated lateral resolution of 1.24$\mu$m with a z-depth of \add{3$\mu$m} interval. The image volume, whose physical size spans approximately \add{1270$\times$930$\times$800 $\mu$m$^3$}, was re-sampled for reconstruction isotropically to a voxel size of 1$\mu$m using bilinear interpolation. The networks were trained using one image sample, and, during training and inference, we used mini-batches with sub-regions of 120-144 pixel. After inference, the sub-regions were stacked back to the original volume space.  In order to provide a reference that confirms the resolution improvement is real, we additionally imaged the sample after physically rotating it by 90 degrees, so that its high resolution XY plane would match the axial \add{XZ} plane of the original volume, sharing the \add{YZ} plane. The reference image volume was then registered on a cell-to-cell level to the input image space using the BigWarp Plugin\cite{Bogovic2016}. While the separately acquired reference is far from a perfect ground-truth image because of its independent imaging condition and potential registration error, \add{it still provides the best available reference as to whether the reconstructed details by the framework match the real physical measurement.}

\begin{figure}[!hbt]
	\centering
	\center{\includegraphics[width =1\textwidth]{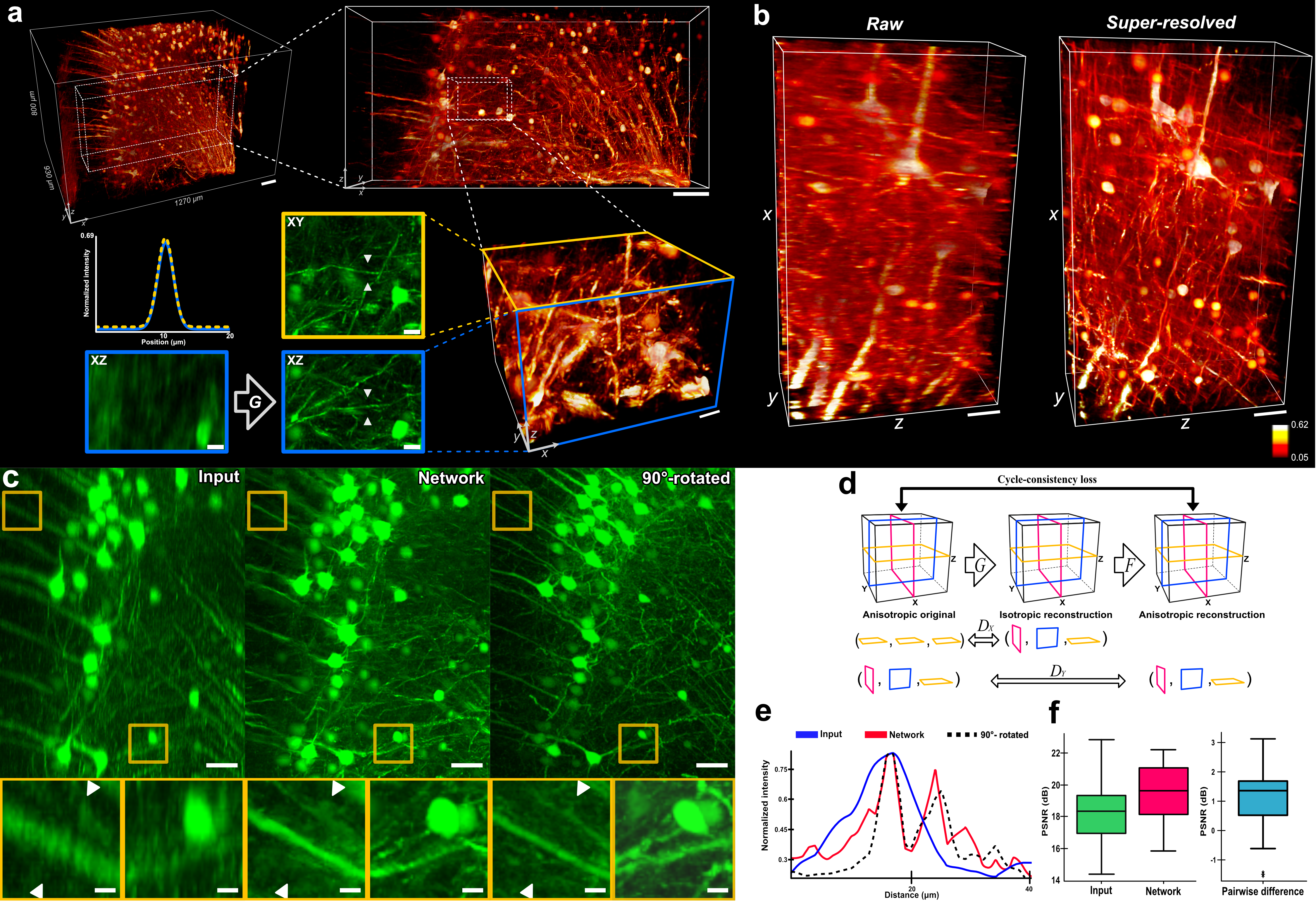}}
	\caption{\linespread{1.1}\footnotesize
	\textbf{Deep learning enables single-volume super-resolution for volumetric CFM.} Cortical regions of a Thy 1-eYFP mouse brain were imaged. \textbf{a}. Isotropic resolution achieved by the proposed method in a brain CFM volume.  Axial images were blindly enhanced by the generative neural network, denoted as $G$ following from \textbf{d}, which is trained and tested on a single CFM image volume that spans 1-2 gigabyte in memory. The resolution improvement in axial planes was global and consistent throughout the image space (also see Supplementary Fig. \ref{fig:snu_overall_sup}). The aligning cross-sectional intensity profiles of a cylindrical dendrite in $XY$ plane (yellow-dotted line) and $XZ$ plane (blue line) indicate near-perfect isotropic resolution, with $XY$ FWHM (full width at half maximum) 2.75 $\mu$m and $XZ$ FWHM of 2.80 $\mu$m, greatly reduced from input $XZ$ FWHM of 8.00 $\mu$m. The cross-sectional intensity profiles were calculated on maximum intensity projection (MIP) images and fitted into 2D Gaussian functions. \textbf{b.} 3D reconstruction of pyramidal neurons in the upper cortical layer before and after restoration. Resolution improvement in axial planes allows more precise and detailed reconstruction of 3D neuronal morphology. Both image volumes are visualized on the same intensity spectrum. \textbf{c.} Image restoration results (\emph{Network}), showing upper cortical regions of the mouse cortex, as MIP images (200 $\mu$m thickness). The results are compared to the original axial imaging (\emph{Input}) and reference lateral imaging (\emph{90\degree-rotated}). Zoomed-in ROIs are marked as yellow boxes. Suppressed or blurred details were recovered in the network output images and match the lateral imaging. The input image and the output image are visualized on the same intensity spectrum. \textbf{d}. Schematic of the framework. The framework learns the data-specific transformation mapping between low-resolution images and high-resolution images by learning to super-resolve axial images and revert the process in a coherent manner. The generative networks, $G$ and $F$, use 3D convolution layers, and the discriminative networks, $D_{X}$ and $D_{Y}$, use 2D convolution layers. Refer to Supplementary Fig. \ref{fig:framework_overview}  and Supplementary Note for more details. \textbf{e}. Cross-sectional intensity profiles from marked lines in zoomed-in ROIs from \textbf{c}. \textbf{f}. PSNR distribution of 32 images as a distance metric to the reference image. Scale bars: \textbf{(a)} 100 $\mu$m, 100 $\mu$m, 20 $\mu$m, 20$\mu$m in the progressively zooming order,  \textbf{(b)} 50 $\mu$m. \textbf{(c)} 50 $\mu$m and 10$\mu$m (ROI) }
	\label{fig:snu_cfm}
\end{figure}

\add{In our test on the CFM image volume, the trained network restored previously highly anisotropic resolution to near-perfect isotropic resolution. In Fig. \ref{fig:snu_cfm}a, we illustrated the resolution improvement by comparing the distance on a resolved axial image and the corresponding distance in the lateral image for an imaged structure that is symmetrical between the lateral and the axial plane: here, a basal dendrite, which is mostly cylindrical. In the example, the difference was nanoscale (~0.05 $\mu$m). As the textural differences between the lateral images and the super-resolved axial images were imperceptible to human eyes, we performed Fourier Spectrum analysis before and after restoration and showed that the frequency information of the output is restored to match that of the lateral imaging (refer to Supplementary Figure \ref{fig:fourier_analysis}).} 

We examined anatomical accuracy of the resolved details by comparing to the reference image, which is labeled as \emph{90\degree-rotated} in Fig. \ref{fig:snu_cfm}c and provides a high resolution match to the original axial image on a submicrometer scale. We noticed that the network was successful not only in translating the axial image texture to the high resolution image domain, but also in recovering previously suppressed details, which were verified by the reference imaging. In comparison to the reference image, the network output accurately enhanced anatomical features of the nervous tissue, consistently throughout the image space (refer to Fig. \ref{fig:snu_cfm}a and Supplementary Fig. \ref{fig:snu_overall_sup}). \add{As shown in Fig. \ref{fig:snu_cfm}b and c, the network allowed for more advanced cytoarchitectonic investigation of the cortical region, as the network managed adaptive recovery of important anatomical features that vary in morphology, density and connectivity across the cortical region. For example, in the upper cortical layers, the previously blurred apical dendrites of pyramidal neurons were resolved, as visualized in fine details by the left-side ROI example of \emph{Network} in Fig. \ref{fig:snu_cfm}c. The network output also revealed previously unseen cortical micro-circuitry by pyramidal neurons and interneurons, as shown in the right-side ROI example of \emph{Network} in Fig. \ref{fig:snu_cfm}c. In the lower cortical layers (refer to Supplementary Fig. \ref{fig:snu_overall_sup}), the network visualized connectivity of extensive axon collaterals, which were previously indiscernible. Such improvements in fine details congregate to reveal important cytoarchitectonic features of the neural circuitry, as shown in Fig. \ref{fig:snu_cfm}b.} The cross-sectional intensity profile (Fig. \ref{fig:snu_cfm}e) illustrates such recovery of suppressed details that were previously blurred in the axial imaging. We noticed that while the network improved the axial resolution, it introduced virtually no discernible distortions or artifacts to the XY (lateral) plane. 

\add{Before we quantify the axial resolution enhancement on a signal level, we needed to minimize the discrepancy in fluorescence emission distribution between the original imaging and the reference imaging. For this reason, we trained and tested the framework separately on another brain sample that is imaged at a higher laser intensity (from 0.3\% power to 3\%) to ensure similar illumination distributions between the original imaging and the reference imaging. The sample size was approximately 870$\times$916$\times$500 $\mu$m$^3$, with lateral resolution of 1.24$\mu$m with a z-depth of \add{4$\mu$m} interval. We then identified 32 non-overlapping ROIs of each 120$\times$120 $\mu$m in the input axial images and the reference lateral images, where identical neuronal structures are distinguishable and detected similarly by fluorescence emission (refer to Supplementary Figure \ref{fig:PNSR_ROI}).} Then we measured and compared the peak signal-to-noise ratio (PSNR) distance of the input and the network output ROIs to the corresponding reference ROIs (Fig. \ref{fig:snu_cfm}e). The network introduced a mean PSNR improvement of 1.15 dB per pair of input ROI versus output ROI. This analysis suggests that the textural details recovered by the network include anatomically accurate features that were more discernible in the lateral imaging. We also examined five cases where the metric improvement by the network output was negative (refer to Supplementary Figure \ref{fig:PNSR_ROI_negative}). We noticed that their metric differences were not indicative of the perceptual accuracy of recovered details and are attributed to differences in fluorescence emission by imaging at a different angle. 

\add{Using the trained network from the PSNR measurement,} we also tested for generalization of the framework by applying to images acquired in different imaging conditions: z-depth sampling rate, intensity of fluorescence light source, and a different sample. We tested on a different brain sample imaged at a lower fluorescence light intensity (lowered from 3\% power to 0.3\%) and a z-depth interval of 3$\mu$m. We also imaged the sample at a perpendicular angle to obtain high-resolution lateral images for reference and registered to the test image on a cellular level. As shown in  Supplementary Fig.~\ref{fig:snu_generalization}, when blindly applied to a new sample, the network output maintained its super-resolution performance, which was consistent through over 800 test images.

\begin{figure}[!hbt]
	\centering
	\center{\includegraphics[width =1\textwidth]{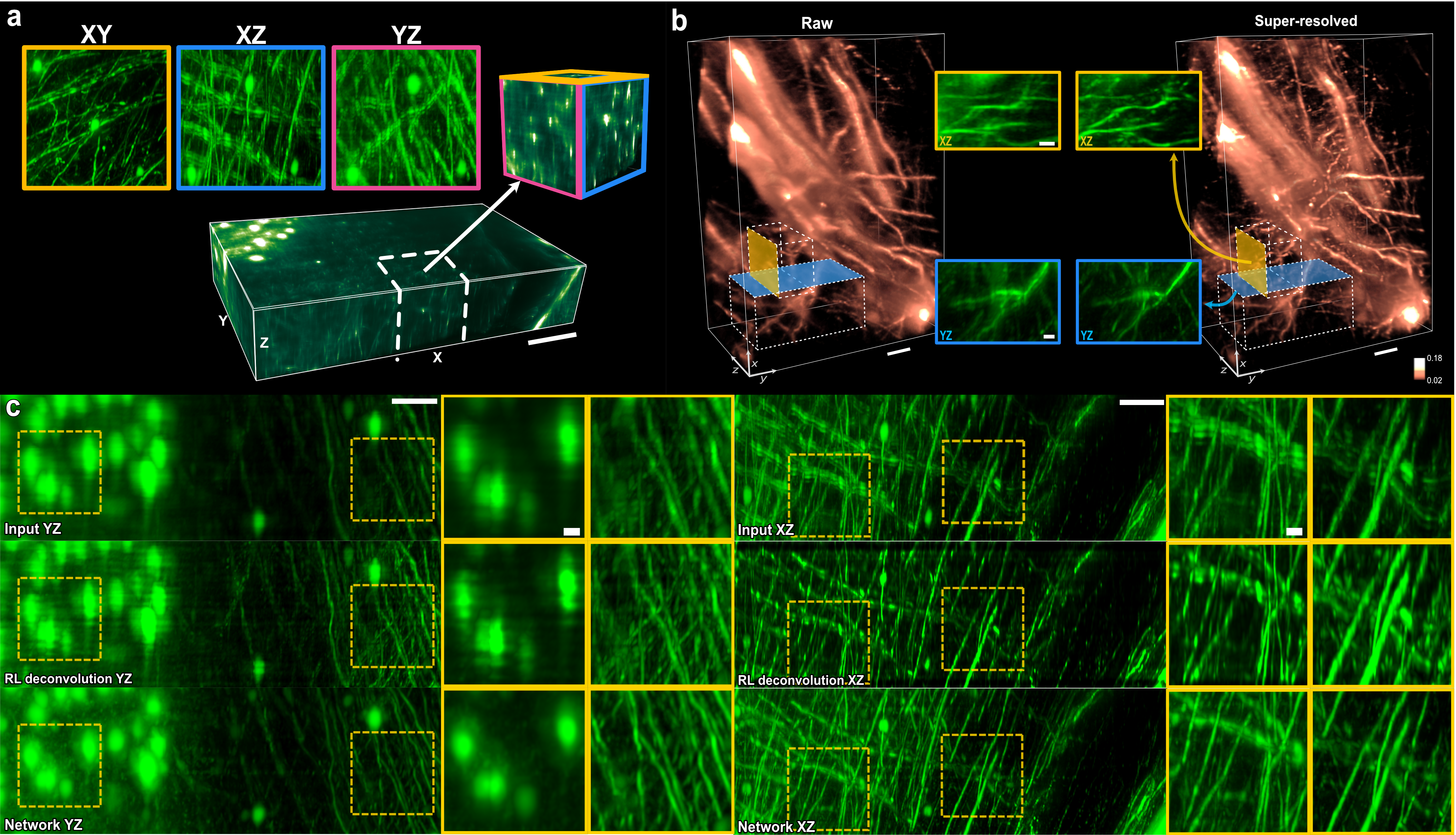}}
	\caption{\linespread{1.1}\footnotesize 
	\textbf{Super-resolution results of the proposed framework in OT-LSM.} \textbf{a}. Comparison of image qualities of orthogonal image planes in OT-LSM, as MIP images of the selected 3D ROI. The design of the OT-LSM system introduces uneven distortions to $XZ$ and $YZ$ plane. \textbf{b}. 3D reconstructions of somata and basal dendrites of pyramidal neurons, with axial MIP images of selected 3D ROIs. The network corrected the sensor drift and also enhanced the contrast between the signals and the background. The resolution enhancement was coherent across $XZ$ plane and $YZ$ plane. The input image and the output image are visualized on the same intensity spectrum.  \textbf{c}.  $YZ$ and $XZ$ MIP images from the input, the deconvolution by Richardson-Lucy algorithm, and the network output. The deconvolution effect was consistent throughout throughout all 8600 slice images of the image volume. Scale bars:   \textbf{(a)} 10 $\mu$m,  \textbf{(b)} 25 $\mu$m,  \textbf{(c)} 50 $\mu$m and 10 $\mu$m (ROI).}
	\label{fig:postech_lsfm}
\end{figure}

\textbf{Resolution enhancement in light-sheet fluorescence microscopy} 
Light-sheet microscopy (LSM) is a specialized microscopic technique for high throughput cellular imaging of large tissue specimens including optically cleared tissues. Current LSM systems have relatively low axial resolutions mainly driven by the optic aberration that is caused by the mismatch of refractive indices between air and immersion medium \cite{ahrens2013whole,glaser2017light}. In particular, open-top LSM (OT-LSM) system \cite{McGorty2015,Mcgorty2017} requires the excitation path and the imaging path to be perpendicular to each other \add{and introduces distortions to the image quality that are uneven between $XZ$ plane and $YZ$ plane}, although this anisotropy could be relaxed by tightly focused excitation \cite{Kim2021}. \add{Therefore, the anisotropy problem for an OT-LSM system usually requires the network to learn two distinct image transformations in each orthogonal plane.}

Here, we again imaged the cortical region of a tissue-cleared mouse brain labeled with Thy 1-eYFP using an OT-LSM, which has the physical size of $\sim$930$\times$930$\times$8600 $\mu$m$^3$. The image was re-sampled for reconstruction isotropically to a voxel size of 0.5$\mu$m using bilinear interpolation. The microscopy system is estimated to have image resolution of $\sim$2$\mu$m laterally and $\sim$4$\mu$m axially, with the z-depth scanning interval of 1$\mu$m. Here, the z-axis is defined as the scanning direction. \add{For the schematics of the OT-LSM system, refer to Supplementary Fig. \ref{fig:postech_system}. As shown in Fig. \ref{fig:postech_lsfm}a, we noticed that the image quality degradation in axial images by our OT-LSM system was nonidentical between $XZ$ and $YZ$ plane, as $XZ$ and $YZ$ images were affected in different degrees by light propagation that aligns with $Y$ axis and vibration from the lateral movement of the cleared imaged sample while scanning.}

\begin{figure}[!hbt]
	\centering
	\centerline{\epsfig{figure=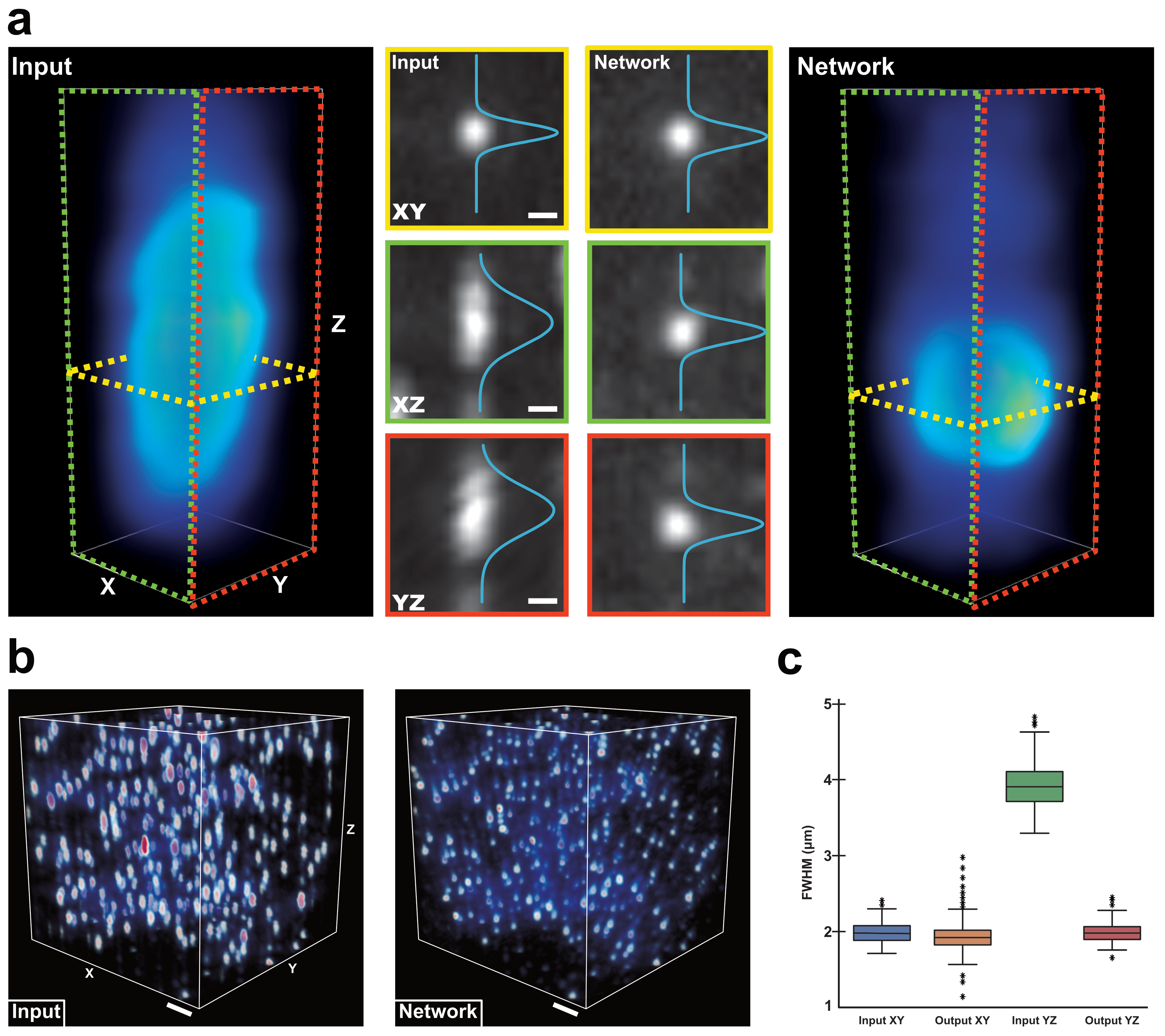, width=1\linewidth}}
	\caption{\linespread{1.1}\footnotesize{ 
	\textbf{PSF Deconvolution by the framework} 
	 0.5$\mu$m fluorescent beads were imaged to experimentally model a PSF of the OT-LSM system. \textbf{a}. An example of PSF deconvolution visualized in 3D and 2D with intensity profiles fit into Gaussian functions. Axial elongation is a common issue in fluorescence microscopy. Our framework resolves it to the originally spherical fluorescent bead. 
	\textbf{b}. A group of fluorescent beads deconvolved in 3D. Fluorescent beads were arbitrarily placed in both the axial and lateral plane.
	\textbf{c}. FWHMs of experimentally measured PSFs in the lateral plane and the axial plane before and after PSF deconvolution. We extracted more than 300 bright spots from the same locations before and applying the method. Each spot was fit into a 2D Gaussian function, where FWHM is calculated. The PSFs in the axial plane are deconvolved to the almost identical resolution as those in the lateral plane. Scale bars: \textbf{(a)} 1 $\mu$m, \textbf{(b)} 20 $\mu$m. }}
	\label{fig:PSF}
\end{figure}

 \add{As shown in Fig. \ref{fig:postech_lsfm}c, the network output showed even balance of enhanced resolution between $XZ$ and $YZ$, while also enhancing the contrast between signals and the background. This improvement enabled more detailed reconstruction of 3D neuronal morphologies, as shown in Fig. \ref{fig:postech_lsfm}b.} To verify the restored details, we deconvolved the image volume using the Richardson-Lucy (RL) deconvolution algorithm \cite{Richardson1972,Lucy1974} based on the PSF model that was experimentally acquired (Fig. \ref{fig:PSF}). The RL deconvolution was performed with Fiji Plugin DeconvolutionLab\cite{Sage2017}, with 10 iterations. In the RL-deconvolved image, we found matching details that were previously suppressed in the input image (Fig. \ref{fig:postech_lsfm}c) and which confirm that our network did not generate any spurious features. \add{Resolution improvement in both axial planes had imperceptible differences in texture and accuracy.} Furthermore, our proposed framework reduced imaging artifacts such as horizontal stripes by the sensor drift in axial imaging, while RL-deconvolved image still contains those (refer to Fig. \ref{fig:postech_lsfm}c and Supplementary Figure \ref{fig:postech_image_artifacts}). We noticed that the improved deconvolution effect by our method was consistent throughout the image volume space.

\textbf{PSF-deconvolution capability} 
To explore further on the image restoration capability of our framework, we tested for the deconvolution capability of an experimentally measured PSF. We imaged 0.5$\mu$m fluorescent beads with OT-LSM in the same imaging condition, with the overall physical size of 360$\times$360$\times$160 $\mu$m$^3$. The image was again re-sampled for reconstruction isotropically to a voxel size of 0.5$\mu$m using bilinear interpolation. The beads were spread arbitrarily, with some of the beads spaced closer to each other. %The learning scheme was the same as in the CFM experiment. 
The results by our framework are shown in Fig.~\ref{fig:PSF}. After the deconvolution, the 2D and 3D reconstruction of the network output indicated almost isotropic resolution, resulting in an almost spherical shape (Fig. \ref{fig:PSF}a). This deconvolution effect was consistent across individual fluorescent beads (Fig. \ref{fig:PSF}b). To quantify the performance of deconvolution, we calculated 2D FWHM values of more than 300 randomly selected bright spots and compared the lateral FWHM values and the axial FWHM values as in before versus after the image restoration. As shown in Fig. \ref{fig:PSF}c, the FWHM distributions of the bright spots in the restored image show an almost identical match to those of the input in the lateral plane. The network output corrected the axial elongation of the PSF, with a mean FHWM of $\sim$3.91 $\mu$m being reduced to $\sim$1.99 $\mu$m, which is very close to a mean FHWM of $\sim$1.98 $\mu$m from the lateral input. The network introduced very little deviation in the lateral plane, with a mean FWHM mismatch of $\sim$0.036 $\mu$m. 

So far, we demonstrated  the effectiveness of our method with CFM and OT-LSM, which involve many dissimilarities between lateral and axial image quality in fluorescence image formation. Accordingly, we expect the framework to be widely applicable to other various forms in the fluorescence microscopy spectrum, as the essential component of the learning does not rely on conditions of an image formation process. 
%Therefore, we conjecture  that the strategy in practice be either to collect microscopy image stacks to the amount that collectively constitutes enough 2D information to reconstruct 3D objects (as in our LSFM example) or to choose a dataset where imaged objects are symmetrical in appearance between the axial plane and the lateral plane (as in our CFM example).

\section*{Discussion}
In summary, we developed a deep-learning based super-resolution technique that enhances axial resolution of conventional microscopy by learning from  high-resolution  lateral images. The strength of our framework comes from taking advantage of learning from unmatched data pairs: it allows the learning of image transformation to be localized to a user-defined 3D unit space and thus to be decoupled from regional variations in image characteristics, such as aberrations or artifacts that arise naturally from the fluorescence imaging process. It also greatly reduces the effort to put into practice as the training of a network requires as little as a single 3D image stack, without a priori knowledge of the image formation process, registration of training data, or separate acquisition of target data. Some combination of those factors is generally considered so far necessary\cite{Belthangady2019} for most conventional deep-learning based super-resolution methods. For this reason, our approach significantly lessens pre-existing difficulty of applying super-resolution to microscopy data.

% Maybe we can discuss about why our model was successful: how it is different from the standard cycleGAN architecture. 

%Super-resolution in biology must be verified even in infinitesimal visual details. 

%%%%%%%%%%%%%%
%%%	METHOD %%%
%%%%%%%%%%%%%%

\section*{METHODS}

\subsection{Sample Preparation and image acquisition}
Tg(Thy1-eYFP)MJrs/J mice were identified by genotyping after heterozygous-mutant mice were bred, and mice were backcrossed onto the C57BL/6 WT background for 10 generations and then maintained at the same animal facility at the Korea Brain Research Institute (KBRI). Mice were housed in groups of 2–5 animals per cage with ad libitum access to standard chow and water in 12/12 light/dark cycle with ``lights-on'' 
at 07:00, at an ambient temperature of 20-22 $^\circ$C and humidity (about 55 \%) through constant air flow. The well‐being of the animals was monitored on a regular basis. All animal procedures followed animal care guidelines approved by the Institutional Animal Care Use Committee (IACUC) of the KBRI
(IACUC-18-00018). In the preparation of the mouse brain slice, the mice were anesthetized by injection with zoletil (30 mg/kg) and xylazine (10 mg/kg body weight) mixture. Mice were perfused with 20 ml of fresh cold PBS and 20 ml of 4 \% PFA solution using a peristaltic pump and whole mouse brain was extracted and fixed 4\% PFA for 1-2 days at 4 $^\circ$C. The fixed mouse brain was sliced coronally in 500 $\mu$m thickness. Then, the brain slices were incubated in the RI matching solution at 36 $^\circ$C for 1 hours for the optical clearing. The proposed method was applied to the images of optically cleared tissues of the brain slices of Thy1-eYFP mice.

For the CFM system, the optically cleared tissue specimens were mounted on the 35-mm coverslip bottom dish and were immersed in RI matching solution during image acquisition using an upright confocal microscope (Nikon C2Si, Japan) with Plan-Apochromat 10 $\times$ lens (NA = 0.5, WD = 5.5 mm). The Z-stacks of optical sections taken at 3 or 4 $\mu$m. 

For the OT-LSM system, we used a recently developed microscopy system\cite{Kim2021} without using tight focusing,  whose design is based on the modification of the water-prism open-top light-sheet microscopy\cite{McGorty2015,Mcgorty2017}. The system includes an ETL (EL-16- 40-TC-VIS-5D-M27, Optotune) as part of the illumination arm for the axial sweeping of the excitation light sheet and an sCMOS camera (ORCA-Flash4.0 V3 Digital CMOS camera, Hamamatsu) in the rolling shutter mode to collect the emission light from the sample. The system uses 10$\times$ air objective lens (MY10X-803, NA 0.28, Mitutoyo) in both the illumination and imaging arms, which point toward the sample surface at +45$^\circ$ and -45$^\circ$, and a custom liquid prism filed with refractive index (RI) matching solution (C match, 1.46 RI, Crayon Technologies, Korea) for the normal light incidence onto the clearing solution. The excitation light source was either 488nm or 532nm CW lasers (Sapphire 488 LP-100, Coherent; LSR532NL-PS-II, JOLOOYO).

\subsection{Image pre-processing}
For the OT-LSM brain images, a median filter of 2 pixel radius was applied to remove the salt-and-pepper noise that arises from fluorescence imaging. All images were normalized to scale affinely between -1 and 1 using percentile-based saturation with bottom and top 0.03\% for the CFM images and 3\% for the OT-LSM images. In both OT-LSM experiments, since the OT-LSM system images a sample at 45\degree or -45\degree, we applied shearing in Y-Z axis as affine transformation to reconstruct a correct sample space.

%%%%%%%%%%%%%%%%%%%%%%%%%%%%%%%%%%%%%%%%%%
\subsection{Cycle-consistent generative adversarial network structure}

To derive our algorithm,
we assume that the high-resolution target space $\Xc$ consists of imaginary 3D image volumes with isotropic resolution  according to a probability measure  $\mu$, while the input space $\Yc$  consists of measured 3-D volumes with anisotropic resolution with poorer axial resolution that follows the  probability measure $\nu$. %We denote such data distributions as $\xb\sim \mu$ and $\yb\sim \nu$.  
According to the recent theory of optimal transport driven cycleGAN\cite{Sim2020}, if we were to transform one image volume in $\Yc$ to $\Xc$, we can solve this problem by transporting the probability distribution $\nu$ to $\mu$ and vice versa in terms of 
statistical distance minimization in $\Xc$ and $\Yc$  simultaneously\cite{Sim2020}, which can be implemented using a cycleGAN.
 In our implementation, it is the role of the discriminative networks to estimate such statistical distances and guide the generative networks to minimize the distances.

Unfortunately, as $\Xc$ consists of imaginary isotropic high resolution volumes,  we cannot measure the statistical distance to $\Xc$ from the generated volumes
directly. 
This technical difficulty can be overcome from the following observation:
since an isotropic resolution is assumed for every 3D volume $\xb \in \Xc$, the planes $XY$, $YZ$ and $XZ$ should have the same resolution as the lateral resolution of the input volume,
%Instead, as each 3-D volume $\xb \in \Xc$ is assume to have isotropic resolution,  its $XY$, $YZ$, and $XZ$ planes should have same
%resolution to the lateral resolution of the input volume, 
i.e. $XY$ plane of the input volume $\yb\in \Yc$.
Accordingly, we  can measure the statistical distance to the imaginary volumes in $\Xc$
by defining the statistical distance as the
sum of the statistical distances in $XY$, $YZ$, and $XZ$ planes using 
%This can be computed using
 the following  least square adversarial loss\cite{Mao2017}:
\begin{align*}
    \mathcal{L}_{\Yc \rightarrow \Xc}(G, D_X) = \mathcal{L}_{\Yc \rightarrow \Xc}(G, D_X^{(1)}) + \mathcal{L}_{\Yc \rightarrow \Xc}(G, D_X^{(2)}) + \mathcal{L}_{\Yc \rightarrow \Xc}(G, D_X^{(3)})
\end{align*}
where 
\begin{align*}
    \mathcal{L}_{\Yc \rightarrow \Xc}(G, D_X^{(1)}) = \mathop{\mathbb{E}}_{\yb\sim \nu}[(D_X^{(1)}(\yb_{xy}) - 1)^2] + \mathop{\mathbb{E}}_{\yb\sim \nu}[(D_X^{(1)}([G(\yb)]_{xy}))^2] \\
    \mathcal{L}_{\Yc \rightarrow \Xc}(G, D_X^{(2)}) = \mathop{\mathbb{E}}_{\yb\sim \nu}[(D_X^{(2)}(\yb_{xy}) - 1)^2] + \mathop{\mathbb{E}}_{\yb\sim \nu}[(D_X^{(2)}([G(\yb)]_{xz}))^2] \\
    \mathcal{L}_{\Yc \rightarrow \Xc}(G, D_X^{(3)}) = \mathop{\mathbb{E}}_{\yb\sim \nu}[(D_X^{(3)}(\yb_{xy}) - 1)^2] + \mathop{\mathbb{E}}_{\yb\sim \nu}[(D_X^{(3)}([G(\yb)]_{yz}))^2]   
\end{align*}
where   the subscripts $xy,yz$ and $xz$ refer to the $xy,yz$ and $xz$ planes, respectively.
Here, $\yb_{xy}$, which is  the $XY$ 2D slice image of $\yb$ image volume,  is used as the
$XY$, $YZ$, and $XZ$ plane references from imaginary isotropic volume distribution $\Xc$ and compared with
the corresponding planes of the deblurred volume $G(\yb)$.
%How do we separately denote GAN loss for generators and discriminators?

On the other hand,  the backward path discriminator group $D_Y$ is trained to minimize the following loss: 
\begin{align*}
    \mathcal{L}_{\Xc \rightarrow \Yc}(F, D_Y) = \mathcal{L}_{\Xc \rightarrow \Yc}(F, D_Y^{(1)}) + \mathcal{L}_{\Xc \rightarrow \Yc}(F, D_Y^{(2)}) + \mathcal{L}_{\Xc \rightarrow \Yc}(F, D_Y^{(3)})
\end{align*}
where
\begin{align*}
    \mathcal{L}_{\Xc \rightarrow \Yc}(F, D_Y^{(1)}) = \mathop{\mathbb{E}}_{\yb\sim \nu}[(D_Y^{(1)}(\yb_{xy}) - 1)^2] + \mathop{\mathbb{E}}_{\xb\sim \mu}[(D_Y^{(1)}([F(\xb)]_{xy}))^2] \\
    \mathcal{L}_{\Xc \rightarrow \Yc}(F, D_Y^{(2)}) = \mathop{\mathbb{E}}_{\yb\sim \nu}[(D_Y^{(2)}(\yb_{xz}) - 1)^2] + \mathop{\mathbb{E}}_{\xb\sim \mu}[(D_Y^{(2)}([F(\xb)]_{xz}))^2] \\
    \mathcal{L}_{\Xc \rightarrow \Yc}(F, D_Y^{(3)}) = \mathop{\mathbb{E}}_{\yb\sim \nu}[(D_Y^{(3)}(\yb_{yz}) - 1)^2] + \mathop{\mathbb{E}}_{\xb\sim \mu}[(D_Y^{(3)}([F(\xb)]_{yz}))^2]   
\end{align*}
so that $XY$, $YZ$, and $XZ$ plane images of the blurred volume $F(\xb)$  follows the distribution of $XY$, $YZ$, and $XZ$ plane images of the input volume $\yb\in \Yc$.

Then, the full objective for the neural network training is given by:
\begin{align*}
\mathcal{L}(G, F, D_X, D_Y) = \mathcal{L}_{\Xc \rightarrow \Yc}(F, D_Y) + \mathcal{L}_{\Yc \rightarrow \Xc}(G, D_X) + \lambda\mathcal{L}_{cyc}(G,F)
\end{align*}
where $\mathcal{L}_{cyc}(G,F)$, as the cycle-consistency loss, is the sum of absolute differences, also known as the L1 loss, between $F(G(\yb))$ and $\yb$. $\lambda$, as the weight of the cycle-consistency loss, is set at 10 in our experiments. 
The objective function of the cycle-consistency-preserving architecture aims to achieve the balance between the generative ability and the discriminative ability of the model as it transforms the image data to the estimated target domain as close as possible, while also preserving the reversibility of the mappings between the domains. The generative versus discriminative balance is achieved by the convergence of the adversarial loss in both paths of the image transformation, as the generative network learns to maximize the loss and the discriminative network, as its adversary, learns to minimize the loss. 
%In our implementation, we adopted
%
%More specifically, the adversarial loss in the forward path by the mapping function as $G : \Yc \mapsto \Xc$ is formulated as follows:
%\begin{align*}
%\mathcal{L}_{\Yc \rightarrow \Xc}(G, D_X) = \mathop{\mathbb{E}}_{\xb\sim \mu}[(D_X(\xb))^2]+\mathop{\mathbb{E}}_{\yb\sim \nu}[(1-D_X(G(\yb)))^2]
%\end{align*}
%Likewise, the adversarial loss for the backward path by the mapping function $F : \Xc \mapsto \Yc$ is formulated as follows: 
%\begin{align*}
%\mathcal{L}_{\Xc \rightarrow \Yc}(F, D_Y) = \mathop{\mathbb{E}}_{\yb\sim \nu}[(D_Y(\yb))^2]+\mathop{\mathbb{E}}_{\yb\sim \nu}[(1-D_Y(F(G(\yb))))^2]
%\end{align*}
%
%
%%the reconstructed input image (e.g. $F(G(x))$) and the real input image, where $\lambda$ regularizes the cycle-consistency loss in the full objective.
%
%
%
%Unlike minimizing the GAN losses by the generative networks, minimization of the GAN loss by the discriminative networks is not symmetric between the forward path and the backward path. In the forward path, the $D_X$ is trained to minimize the following loss. 
%

The resulting architecture consists of two deep-layered generative networks, respectively in the forward path and the backward path, and six discriminative networks, in two groups also respectively in the forward path and the backward path. The schematic is illustrated in Fig. \ref{fig:snu_cfm}d and Supplementary Figure \ref{fig:framework_overview}.

Our generative network structure in the forward path, $G$, is based on the 3D U-Net architecture\cite{Ronneberger2015}, which  consists of the downsampling path, the bottom layer, the upsampling path, and the output layer. 
The schematic of this network architecture is illustrated in Supplementary Fig.~\ref{fig:network_designs}a, whose detailed explanation is given in Supplementary Note. 
On the other hand, the generative network architecture in the blurring path, $F$, is adjustable and replaceable based on how well the generative network can emulate the blurring or downsampling process in the backward path. We searched for an optimal choice empirically between the 3D U-net architecture and the deep linear generator\cite{Bellkligler2019} without the downsampling step (refer to Supplementary Figure \ref{fig:network_designs}b). For the CFM brain images and the OT-LSM fluorescent bead images, we chose deep linear generator as $F$, while 3D U-Net is used for the OT-LSM brain images.
%  However, we did not find significant differences in performance for choosing either 3D U-net or deep linear generator, although
%  training with the deep linear generator converges faster.
  Note that the kernel sizes in the deep linear generator vary depending on depths of the convolution layers, as shown in Supplementary Figure \ref{fig:network_designs}b.

\subsection{Algorithm implementation and training}
Before the training phase for the OT-LSM images, we diced the entire volume into sub-regions of 200-250 pixel with overlapping adjacent regions of 20-50 pixels. The number of sub-regions is $\sim$3000 for the brain image data and $\sim$580 for the fluorescent bead image data. Then we randomly cropped a region for batch training per iteration and flipped it in 3D on a randomly chosen axis as a data augmentation technique. The crop size was 132$\times$132$\times$132 for the brain images and 100$\times$100$\times$100 for the fluorescent bead images. 

While the axial resolution in OT-LSM differs between $XZ$ and $YZ$ plane because of the illumination path in alignment with $YZ$ axis, the axial resolution from the CFM imaging is consistent across the $XY$ plane. For this reason, for the CFM images, we loaded the whole image volume (1-2 Gigabytes) in memory and randomly rotated along the $Z$-axis as a data augmentation technique. Then we also randomly cropped a region and flipped on a randomly chosen axis per iteration. For this reason, the networks were trained with one whole image volume with its training progress marked in iterations instead of in epochs. The crop size is set as 144$\times$144$\times$144. During the inference phase in all experiments, the crop size is set as 120$\times$120$\times$120 with overlapping regions of 30 pixels, and we cropped out the borders (20 pixels) of each output sub-region to remove weak signals near the borders before assembling back to the original volume space.  In all experiments, the batch size per iteration is set as 1.

In our 3D U-net generative networks, all 3D convolution layers have the kernel size of 3, the stride of 1 with the padding size of 1, and all transposed convolution layers have the kernel size of 2, the stride of 2, and no padding. In the deep linear generative networks, the six convolution layers have the kernel sizes of [7,5,3,1,1,1] in turn with the stride of 1 and the padding sizes of [3,2,1,0,0,0]. In the discriminative networks, the convolution layers have the kernel size of 4, the stride of 2, and the padding size of 1.

In all experiments, all the learning networks were initialized using Kaiming initialization\cite{He2015} and optimized using the adaptive moment estimation (Adam) optimizer\cite{Kingma2017} with a starting learning rate $1\times10^{-4}$. For the CFM images and the OT-LSM brain images, the training was carried out on a desktop computer with GeForce RTX 3090 graphics card (Nvidia) and Intel(R) Core(TM) i7-8700K CPU @ 3.70GHz. 
%The final model for the CFM images was selected at 84,000th iteration. %, which took $\sim$159 hours to train. 
%The final model for the CFM images in the additional experiment was selected at 72,000th iteration, %, which took ~164 hours to train. 
%The final model for the OT-LSM brain images was selected at 37th epoch. %, which took $\sim$174 hours to train.
 For the OT-LSM fluorescent bead images, the training was carried out on a desktop computer with GeForce GTX 1080 Ti graphics card (Nvidia) and Intel(R) Core(TM) i7-8086K CPU @ 4.00GHz.  %The final model was selected at 68th epoch, which took $\sim$17 hours to train. 

 \begin{addendum}
{\color{black} \item[Correspondence] Correspondence and requests for materials should be addressed to Jong Chul Ye. (email: jong.ye@kaist.ac.kr).}
 \item This research was  supported by  National Research Foundation of Korea (Grant NRF-2020R1A2B5B03001980 and NRF-2017M3C7A1047904).

\item[Author Contributions] J.C.Y. supervised the project in conception and discussion and wrote the manuscript. H.P. conceived and implemented the research. M.N. generated the CFM imaging data. S.C. supervised the CFM imaging. B.K. and S.P. generated the OT-LSM imaging data. K.K. supervised the OT-LSM imaging. H.P. wrote the manuscript.

 \item[Competing Interests] The authors declare that they have no competing financial interests.

%\item[Data and Code Availability] 
% Data or codes are not public yet. Should we make them public?

\end{addendum}

%\clearpage

%\newpage
\bibliographystyle{naturemag}
\bibliography{ref}

\clearpage
%\appendix

%%\newcommand{\beginsupplement}{%
%        \setcounter{table}{0}
%        \renewcommand{\thetable}{S\arabic{table}}%
%        \setcounter{figure}{0}
%        \renewcommand{\thefigure}{S\arabic{figure}}%
%%     }
%     \beginsupplement
\setcounter{equation}{0}
\setcounter{figure}{0}
\setcounter{table}{0}
\setcounter{page}{1}
\renewcommand{\figurename}{Supplementary Figure}
\renewcommand{\tablename}{Supplementary Table}
\renewcommand{\thesection}{S\arabic{section}}
\renewcommand{\thetable}{S\arabic{table}}
\renewcommand{\thefigure}{S\arabic{figure}}

%\section{Supplementary Note}

\section*{\Huge Supplementary Figures}

\begin{figure}[!ht]
\center{\includegraphics[width =0.9\textwidth]{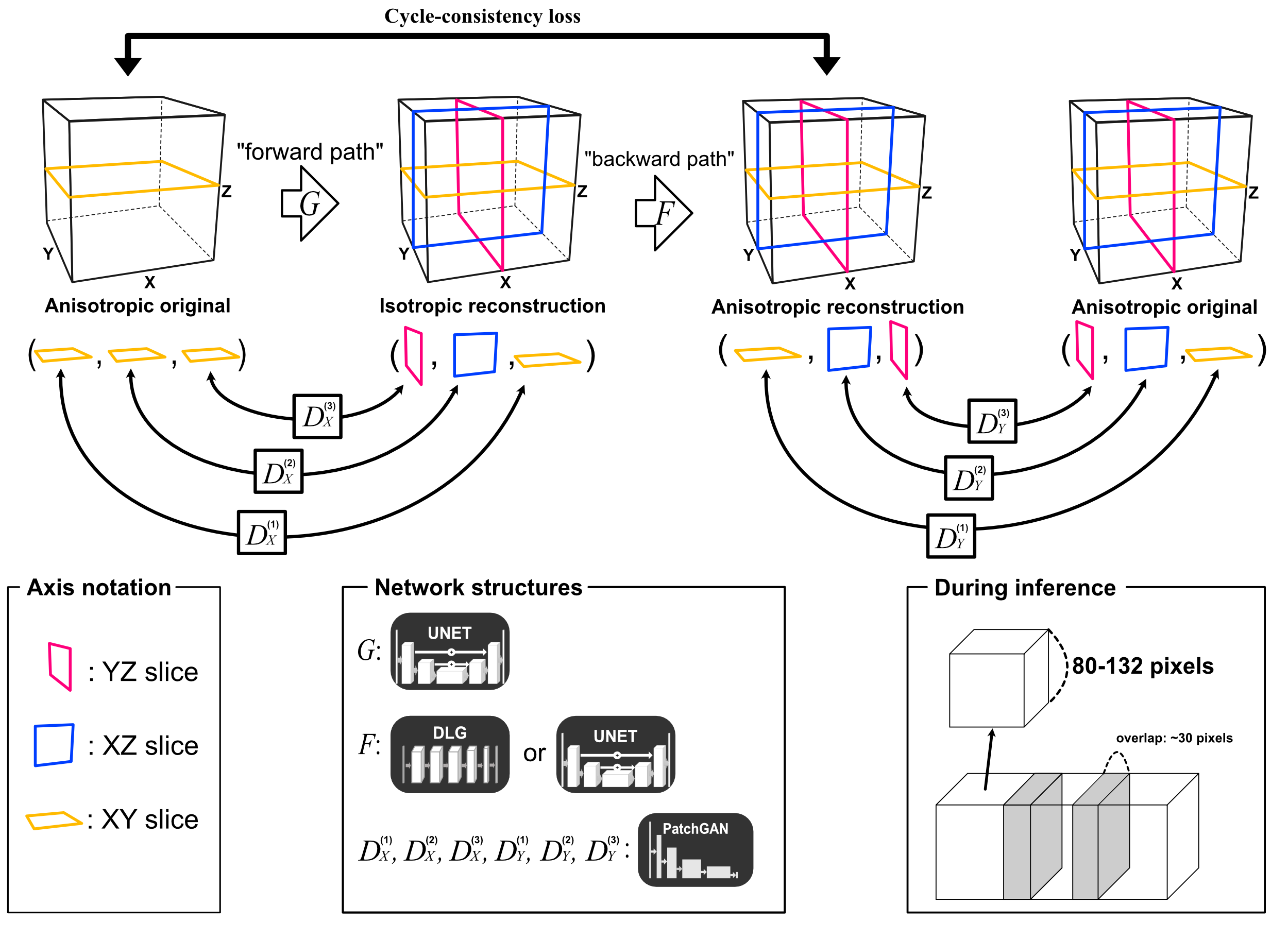}}
\caption{\linespread{1.2}\footnotesize{\textbf{Detailed overview of the framework.} The framework employs two generative networks, $G$ and $F$, and six discriminitive networks ($D_X^{(1)}$, $D_X^{(2)}$, $D_X^{(3)}$, $D_Y^{(1)}$, $D_Y^{(2)}$, $D_Y^{(3)}$). $G$ in the super-resolving path is a 3D U-net, and $F$ in the reverting path can be either a 3D U-net (labeled as UNET) or a deep linear generator (labeled as DLG). $D_X$'s and $D_Y$'s are all PatchGAN discriminators. During the inference phase, the trained $G$ is applied to sub-regions with overlapping neighboring regions iteratively.}}
\label{fig:framework_overview}
\end{figure}

\begin{figure}[!ht]
\center{\includegraphics[width =1\textwidth]{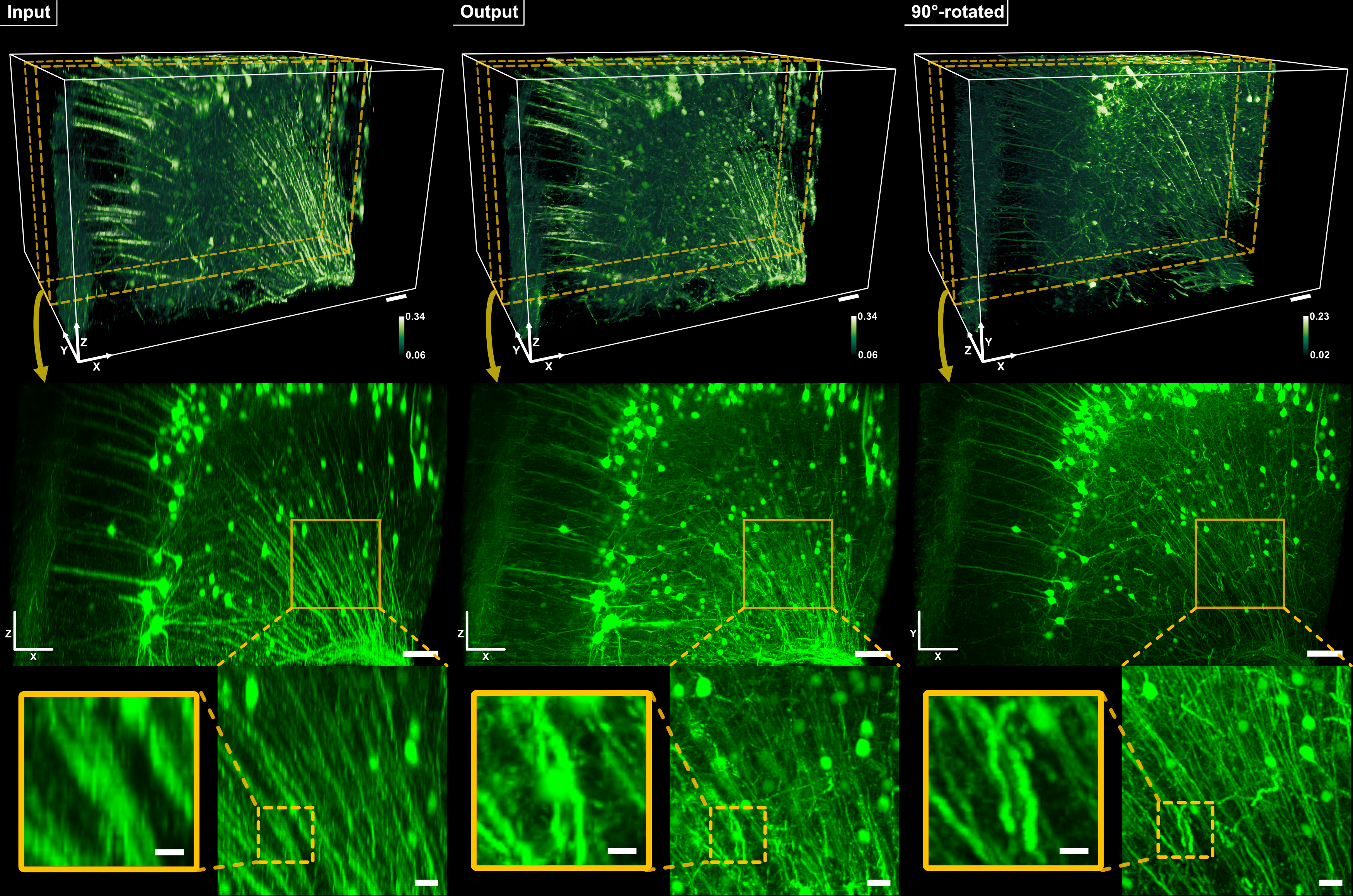}}
\caption{\linespread{1.2}\footnotesize{\textbf{Large-scale enhancement of axial resolution compared to the reference lateral imaging, shown with progressively zoomed-in ROIs}. The images in the second row and the third row are maximum-intensity-projection images of 200 $\mu$m thickness. Scale bar: 100 $\mu$m, 100 $\mu$m, 25 $\mu$m, 10 $\mu$m, in the order of progressive zooming-in. }}
\label{fig:snu_overall_sup}
\end{figure}

\begin{figure}[!ht]
\center{\includegraphics[width =0.8\textwidth]{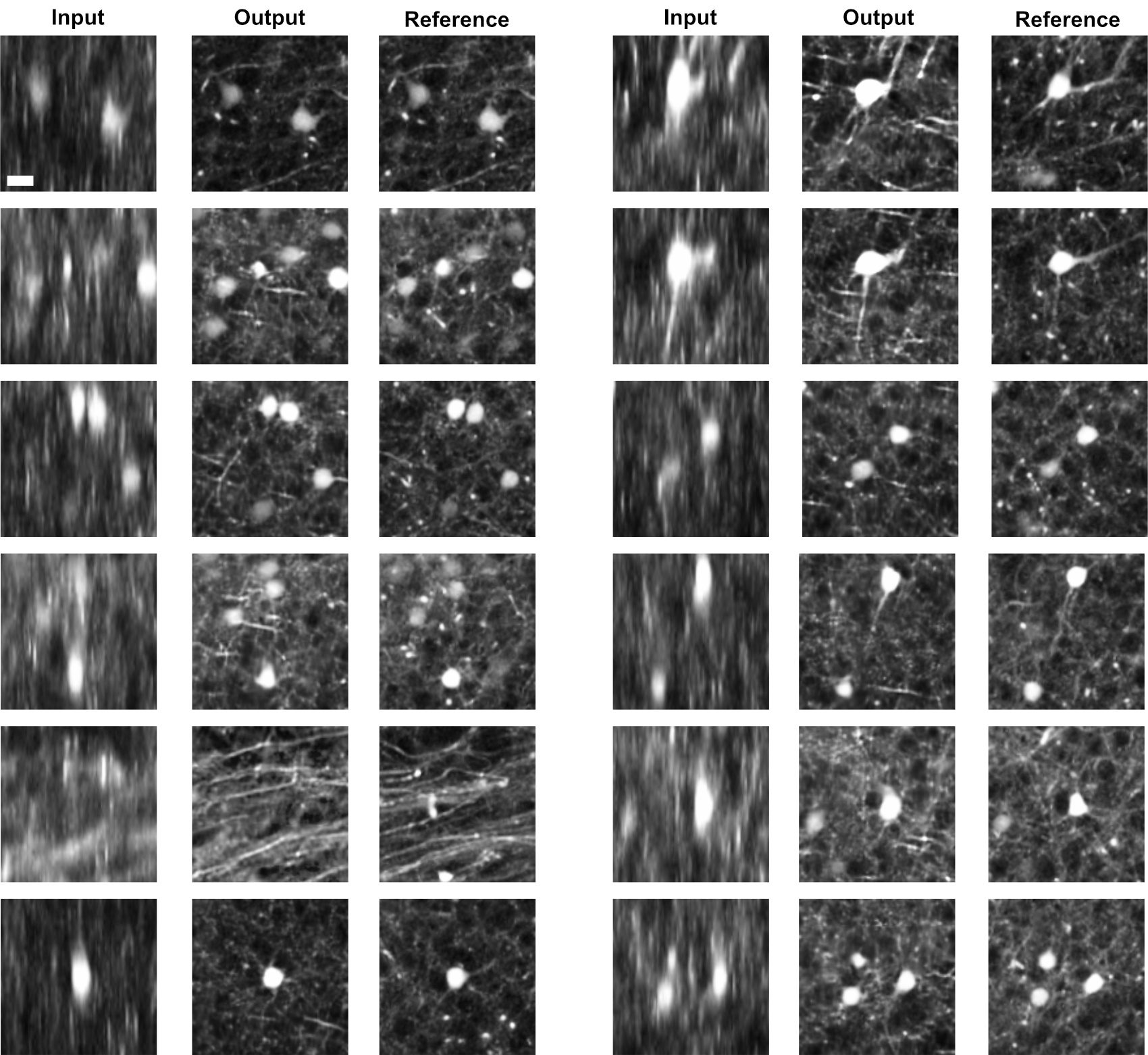}}
\caption{\linespread{1.2}\footnotesize{\textbf{Randomly selected examples out of 32 ROIs for PSNR calculation.} Scale bar: 20 $\mu$m}}
\label{fig:PNSR_ROI}
\end{figure}

\begin{figure}[!ht]
\center{\includegraphics[width =0.6\textwidth]{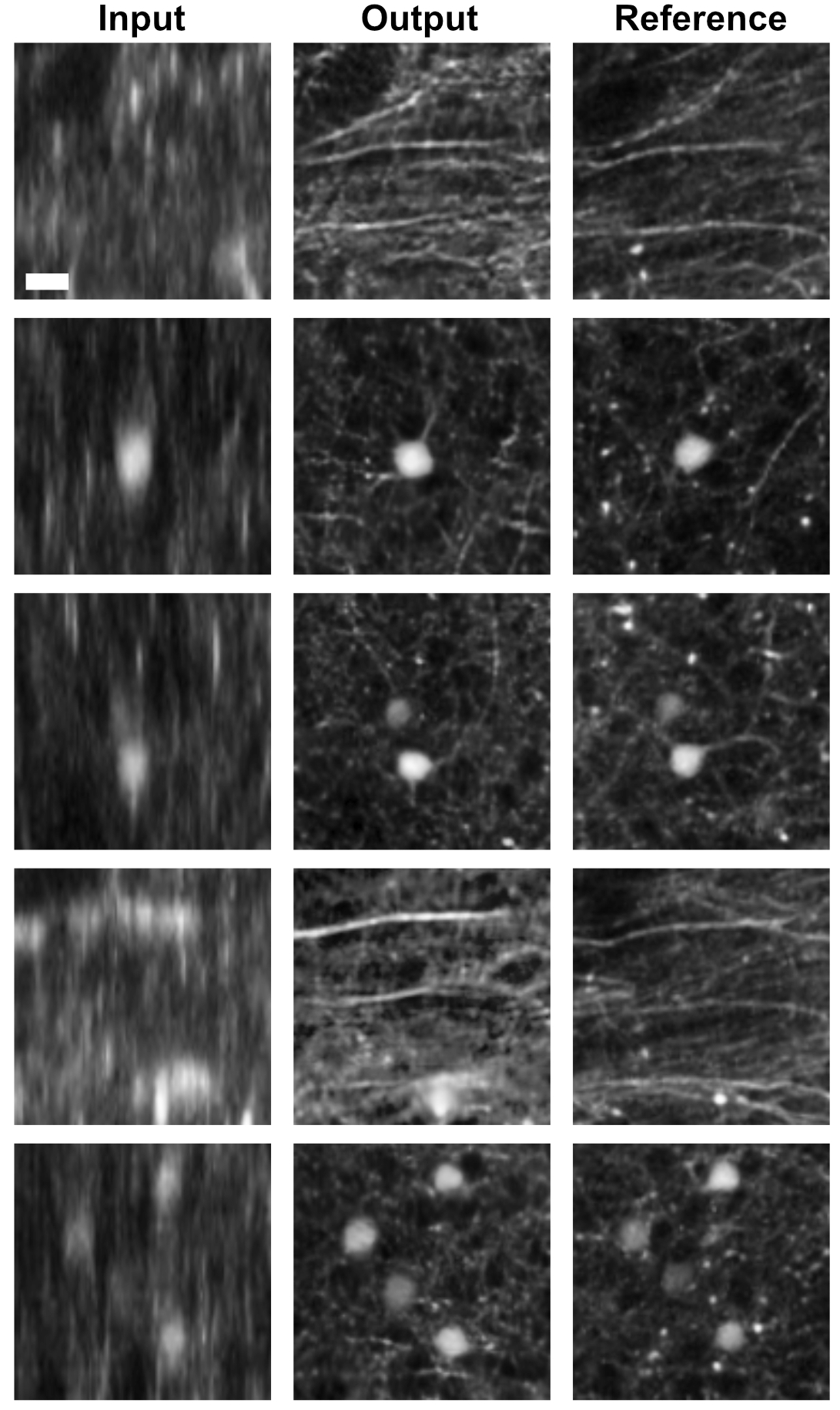}}
\caption{\linespread{1.2}\footnotesize{\textbf{ ROIs with negative PSNR metrics.} Scale bar: 20 $\mu$m}}
\label{fig:PNSR_ROI_negative}
\end{figure}

\begin{figure}[!ht]
\center{\includegraphics[width = 0.8\textwidth]{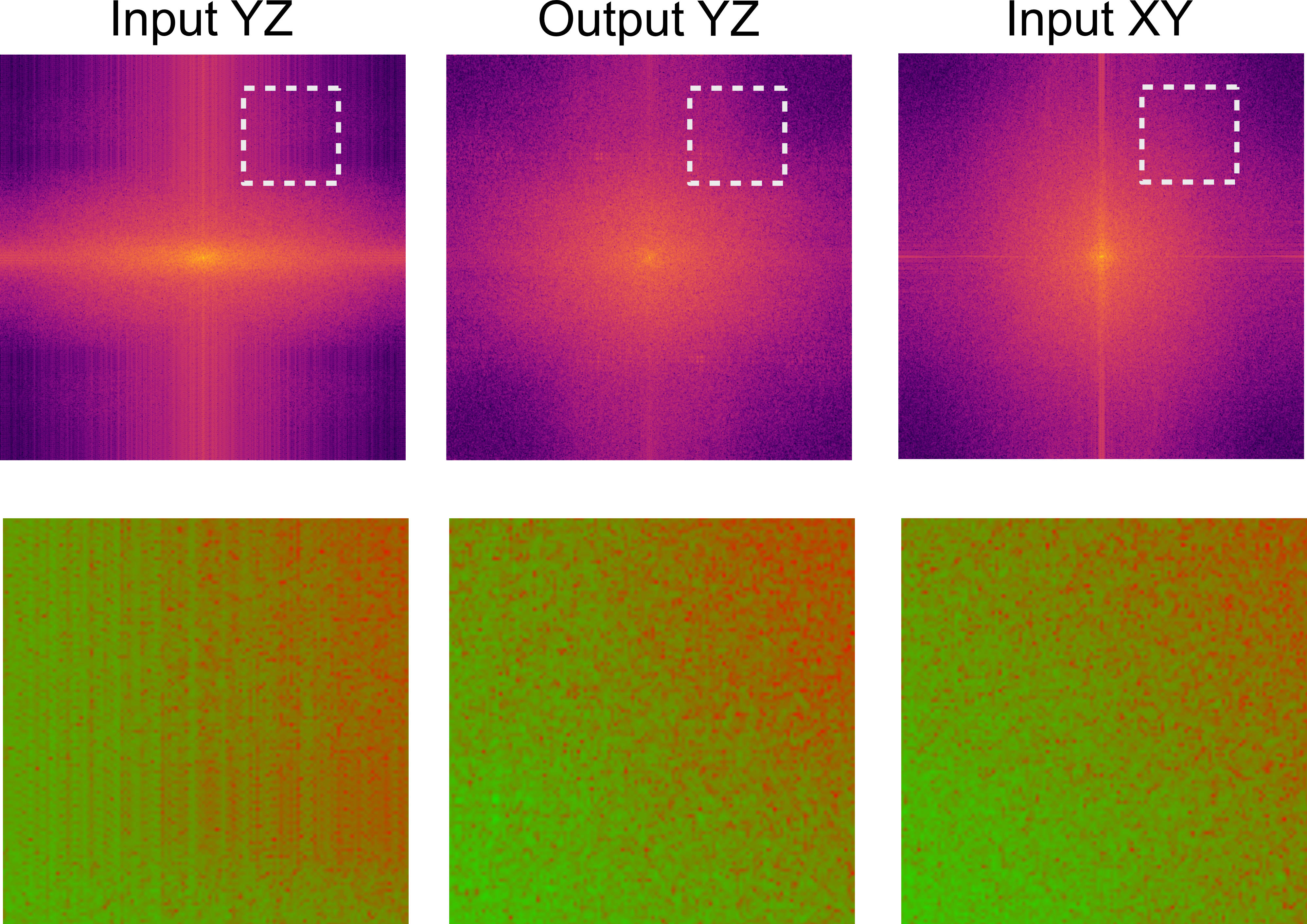}}
\caption{\linespread{1.2}\footnotesize \textbf{Fourier spectrum analysis of CFM images with zoomed-in ROIs that are visualized on a different color-map.} The frequency profile of output YZ illustrates the restoration of the frequency information. The proposed method not only approximates the frequency profile of input XY (i.e. lateral plane image), also restores the loss of information which is visualized as vertical stripes in the frequency profile of input YZ.}
\label{fig:fourier_analysis}
\end{figure}

\begin{figure}[!b]
	\centering
	\center{\includegraphics[width =1\textwidth]{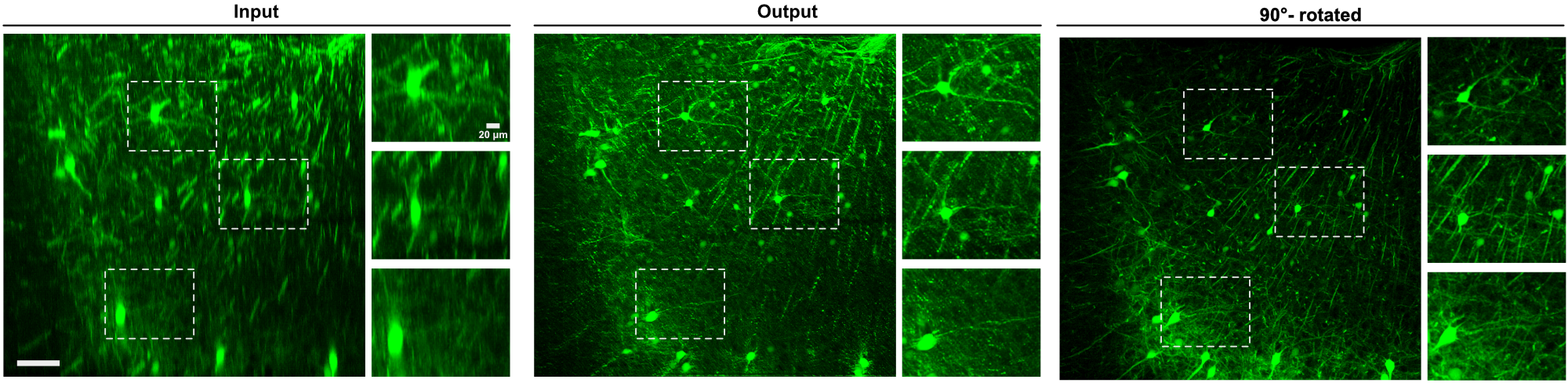}}
	\caption{\linespread{1.2}\footnotesize 
	\textbf{Generalization results of the trained network in CFM.} visualized as maximum intensity projection of 30 $\mu$m thickness. We tested the generalization capability of the framework by applying the trained network to a image of a different sample in different imaging conditions: different sampling z-depth rate and the excitation light power intensity. To verify the results, we also imaged the sample at a perpendicular angle. As before, the network deconvolves the image and enhances the details that are shown to be biologically true. Scale bar: 20 $\mu$ m (ROIs).
	}
	\label{fig:snu_generalization}
\end{figure}

\begin{figure}[!ht]
\center{\includegraphics[width =0.5\textwidth]{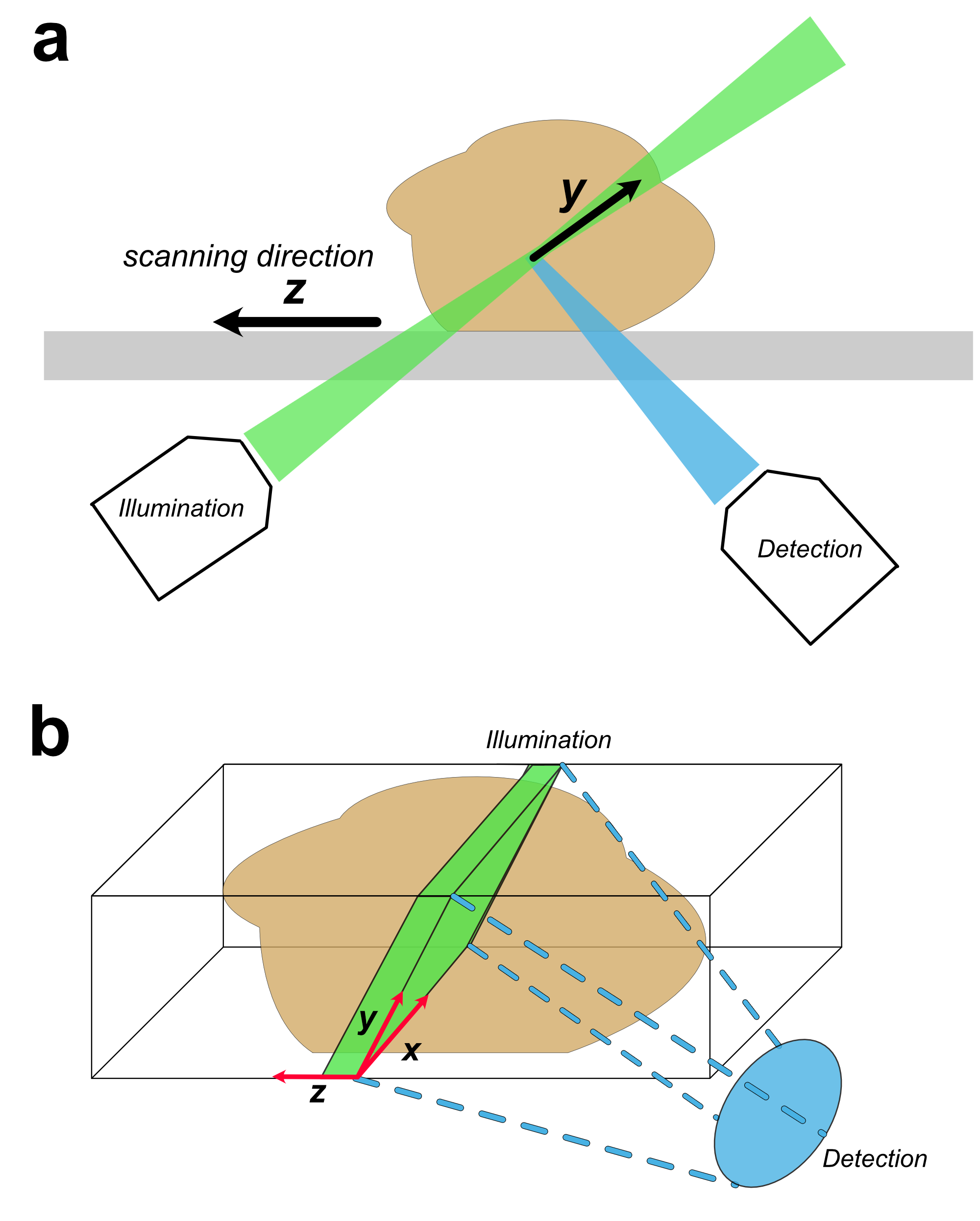}}
\caption{\linespread{1.2}\footnotesize \textbf{Schematic of the OT-LSM system.} \textbf{a.} Lateral view of the imaging system. \textbf{b.} 3D view of the image space with the imaging set-up.}
\label{fig:postech_system}
\end{figure}

\begin{figure}[!ht]
\center{\includegraphics[width =0.8\textwidth]{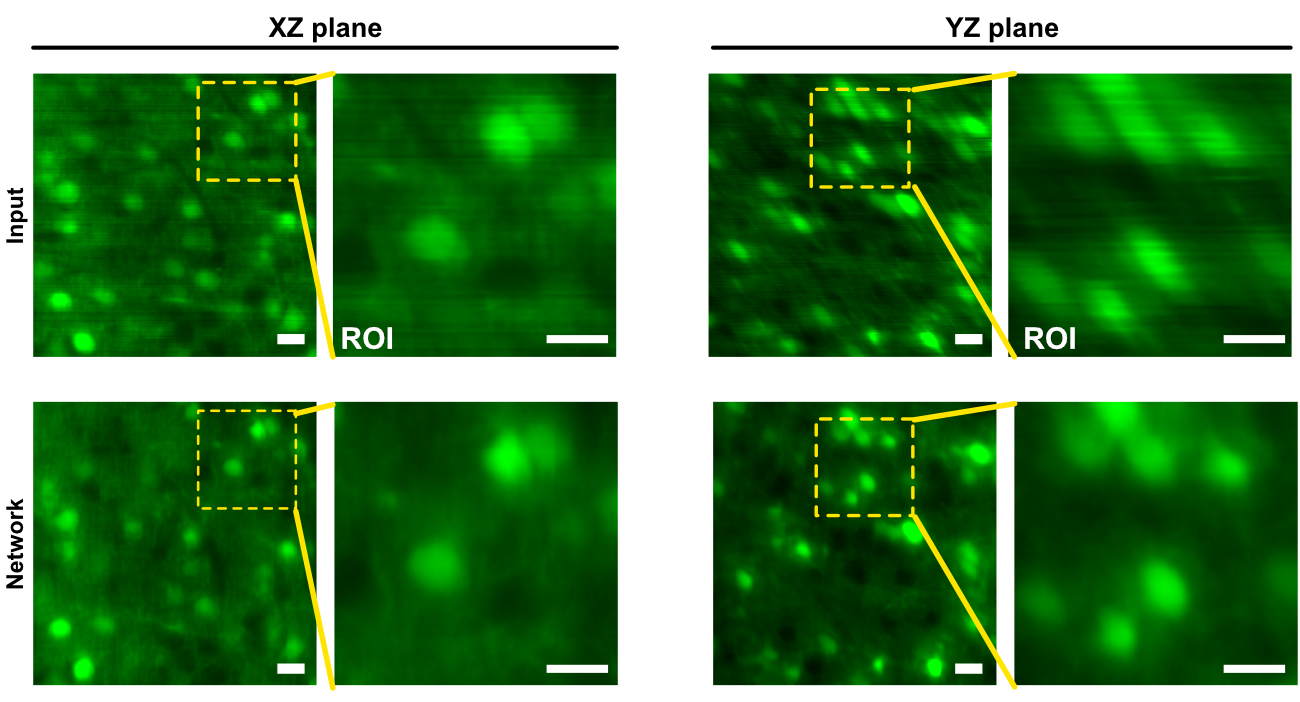}}
\caption{\linespread{1.2}\footnotesize \textbf{Example of the network removing imaging artifacts by sensor drifts.} Scale bars: 25 $\mu$m}
\label{fig:postech_image_artifacts}
\end{figure}

\begin{figure}[!ht]
\center{\includegraphics[width =0.9\textwidth]{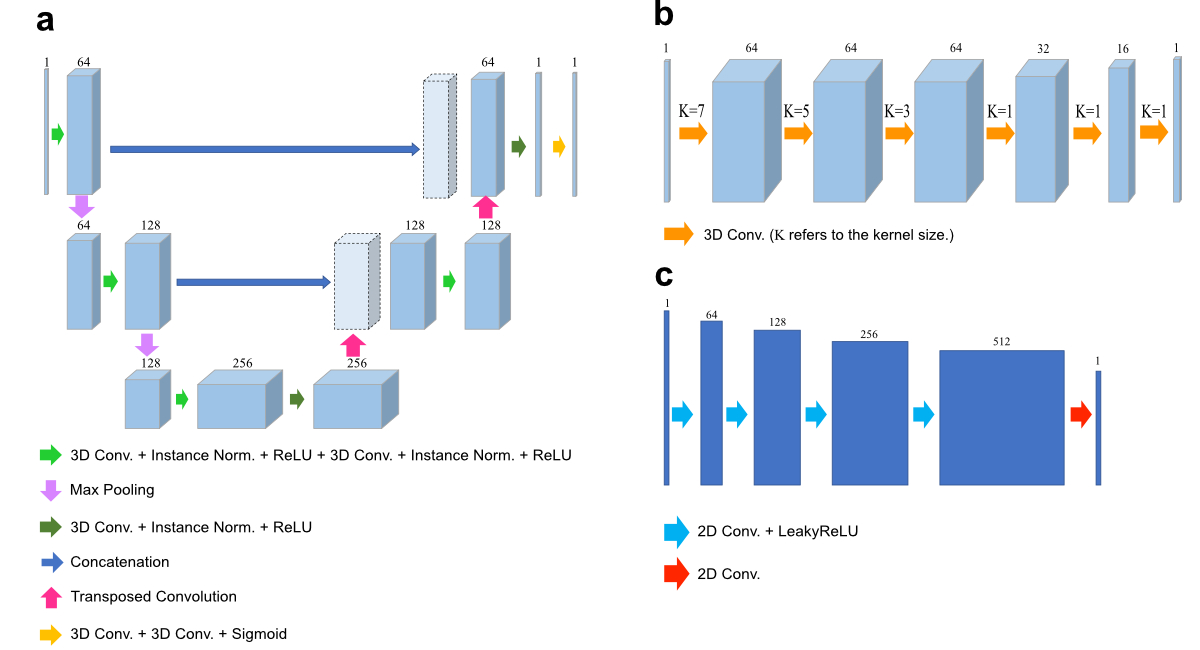}}
\caption{\linespread{1.2}\footnotesize{\textbf{Network designs for the generative networks and the discriminative networks.} \textbf{a.} 3D U-Net architecture for the generative network (for super-resolving path). The kernel size for convolutions is set as 3. \textbf{b.} 3D deep linear generator architecture for the generative network (for blurring path). \textbf{c.} 2D patch-GAN architecture for the discriminative network. The kernel size for convolutions is set as 4.}}
\label{fig:network_designs}
\end{figure}

%\begin{figure}[!ht]
%\center{\includegraphics[width =0.6\textwidth]{figs/manifold_space.jpg}}
%\caption{\footnotesize \textbf{theoretical background of formulating a cycle-consistency-loss preserving network structure as a solution for an optimal transport problem.}}
%\label{fig:manifold_space}
%\end{figure}

\clearpage

\section*{\Huge Supplementary Note}

\subsection{Generative network}

Our generative network structure for the super-resolving path is based on the 3D U-Net architecture\cite{Ronneberger2015}, as illustrated in Supplementary Fig.~\ref{fig:network_designs}a.
The generator is implemented as and encoder-to-decoder architecture and consists of the downsampling path, the bottom layer, the upsampling path, and the output layer. 
Specifically, the downsampling path consists of the repetition of the following block: 
\begin{align*}
\fb_{k}' = \textrm{ReLU} \{[\textrm{Norm}(\textrm{Conv}&\{\textrm{ReLU}[\textrm{Norm}(\textrm{Conv}\{\fb_{k-1}\})]\}\})] \\ 
\fb_{k} &= \textrm{Maxpool}[\fb_{k}'], \quad
k = 1,2
\end{align*}
where 
$\fb_{k}$ represents the output 3D feature tensor of the $k$th down-sampling block, and $\fb_0$ is the input 3D volume. ReLU[] is the the rectified linear unit activation function with a slope of $\alpha$=1, Norm() is the instance normalization\cite{Ulyanov2016}, Conv\{\} is the convolution operation, and the Maxpool[] is the max pooling operation. The bottleneck layer is as follows:
\begin{align*}
	\gb_{0} = \textrm{ReLU}[\textrm{Conv}\{\textrm{ReLU}[\textrm{Conv}\{\textrm{ReLU}[\textrm{Conv}\{\fb_{2}\}]\}]\}]
\end{align*}
where
$\gb_{0}$ is the output of the bottom layer. The up-sampling path consists of the repetition of the following block: 
\begin{align*}
\gb_{k}' = \textrm{ReLU}[\textrm{Norm}(\textrm{Conv}\{&\textrm{ReLU}[\textrm{Norm}(\textrm{Conv}\{\textrm{Concat}[\gb_{k-1}, \fb_{k-1}']\})]\})] \\
%\gb_{1} = \textrm{TrConv}[\gb_{1}'] \\
%y_{2}' = ReLU[Norm(Conv\{&ReLU[
\gb_{k} &= \textrm{TrConv}[\gb_{k}'] ,
\quad k =  1,2
\end{align*}
where
$\gb_{k}$ is the output of the $k$th upsampling block. Concat[] is the concatenation operation, and the TrConv \{\} is the transposed convolution. The last output layer is as follows:
\begin{align*}
	\yb = \textrm{Sigmoid}[\textrm{Conv}\{\textrm{Conv}[\gb_{2}]\}] 
\end{align*}
where $\yb$ is an output 3D volume.

The generative network architecture in the backward path is adjustable and replaceable based on how well the generative network can emulate the blurring or downsampling process in the backward path. We searched for an optimal choice empirically between the 3D U-net architecture (refer to Supplementary Figure \ref{fig:network_designs}a) and the deep linear generator without the downsampling step (refer to Supplementary Figure \ref{fig:network_designs}b). 
%In the CFM experiment and the OTAS-SLM experiment with the measured PSF, we chose deep linear generator as $F$. 
 The kernel sizes in the deep linear generator vary depending on depths of the convolution layers, as shown in Supplementary Figure \ref{fig:network_designs}b.
%\begin{align*}
%	\wb_{k}' = \textrm{LeakyReLU}\{\textrm{Conv}[\gb_{k-1}]\} \\
%	\quad k =  1,2,3,4 \\
%	\wb_{k} = \textrm{Conv}[\wb_{4}']
%\end{align*}

%In the OTAS-SLM experiment with the brain sample, we noticed that the model with the U-Net architecture as $F$ generated more realistic visuals, and the design is the same as $G$ in this case. 

\subsection{Discriminative network structure}

As the inputs to the discriminator networks are  $XY, YZ$, and $ZX$ plane images, 
we adopted the discriminative network structure from 2D patchGAN\cite{Isola2017} for our discriminator networks. The detailed schematic is illustrated in Supplementary Figure \ref{fig:network_designs}c. The patchGAN consists of multiple convolution blocks that allow the discriminator module to judge an input image based on different scales of patches.
\begin{align*}
	\vb_{k} = \textrm{LReLU}[\textrm{Norm}&(\textrm{Con})\{\vb_{k-1}\})],\quad
	k = 1,2,3,4 
\end{align*}
where
$\vb_{0}$ is the input 2D image, either real or fake as generated by the generator network. Norm() is the instance normalization. LReLU is the leaky rectified linear unit activation function with a slope of $\alpha=0.2$. The last layer is a convolution layer that generates a single channel prediction map.

\subsection{Definition of PSNR}
The PSNR metric in this study is calculated as follows:
$$PSNR = 10\log_{10}\left(\frac{N_xN_y\max(r(x,y))^{2}}{\sum_{x=0}^{N_x-1}\sum_{x=0}^{N_y-1}\Bigl[r(x,y)- t(x,y)\Bigr]^{2}}\right)$$
where $r(x,y)$ is the reference image, $t(x,y)$ is the reconstructed image, and both images have the dimensions of $N_x$ and $N_y$.

\end{document}